\documentclass[10pt]{article}
\usepackage{freya}
\usepackage{mathtools}
\usepackage{multirow}\usepackage{makecell}\usepackage{subcaption}\usepackage{wrapfig}
\usepackage{pifont}\usepackage{xspace}\usepackage{url}\usepackage{nicefrac}
\usepackage{float}
\usepackage{enumitem}\usepackage{threeparttable}
\usepackage{cleveref}

\newcommand{\modelname}[0]{FreyaTTS}

\def\huggingface{\raisebox{-1.5pt}{\includegraphics[height=1.05em]{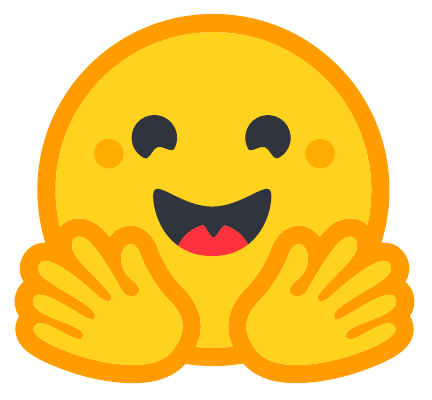}}}
\def\github{\raisebox{-1.5pt}{\includegraphics[height=1.05em]{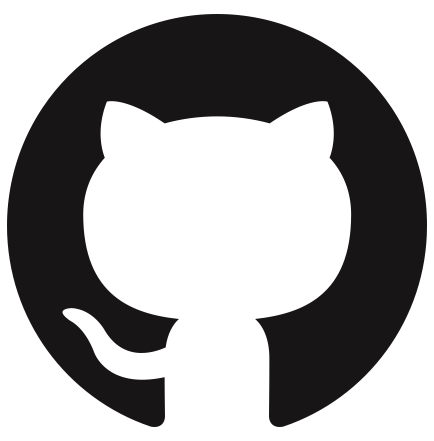}}}

\title{FreyaTTS: A Compact Tokenizer-Free Flow-Matching Transformer for Turkish-First Speech Synthesis}
\author{Freya Team}
\freyarunning{FreyaTTS}
\freyalogo{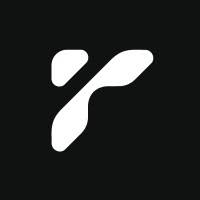}
\freyalinks{%
  \begin{tabular}{@{}rl@{}}
    \github~Project: & \url{https://github.com/freyavoiceai/FreyaTTS} \\
    \huggingface~Model: & \url{https://huggingface.co/freyavoice/Freya-TTS} 
  \end{tabular}}

\begin{document}
\newcommand{\OneSftWer}{8.0}
\newcommand{\OneSftCer}{3.0}
\newcommand{\OnePiperWer}{4.4}
\newcommand{\OneMmsWer}{6.8}
\newcommand{\OneXttsWer}{11.1}
\newcommand{\OneFfiveWer}{24.3}
\newcommand{\OneXttsCer}{3.9}
\newcommand{\OneFfiveCer}{10.9}
\newcommand{\OneNsys}{7}
\newcommand{\OneNitems}{495}
\newcommand{\OneSftWerRank}{3}
\newcommand{\OneSftCerRank}{3}
\newcommand{\OneNsysTotal}{7}

\maketitle

\begin{abstract}
We introduce \modelname{}, a compact, tokenizer-free, Turkish-first text-to-speech model designed for highly reliable and efficient conversational synthesis.
\modelname{} is a $183.2$\,M-parameter non-autoregressive conditional flow-matching Diffusion Transformer (DiT) that operates in the continuous-latent space of the frozen AudioVAE2~\citep{zhou2026voxcpm2}, reused unmodified ($16$\,kHz encode, $48$\,kHz decode); holding the codec fixed lets the model devote all of its capacity to the text-to-latent map while inheriting $48$\,kHz reconstruction for free.
We advance the framework across three key dimensions: 
(i)~\textbf{Rule-Free End-to-End Modeling}, by driving generation end-to-end from a $92$-symbol Turkish character vocabulary with no phonemizer, grapheme-to-phoneme frontend, or discrete speech tokenizer, so that agglutinative morphology, vowel harmony, and the pronunciation of in-context words and acronyms are learned directly from audio (digit strings are expanded to their spoken form at the text frontend);
(ii)~\textbf{Non-Autoregressive Parallel Denoising}, which avoids the left-to-right error accumulation of autoregressive decoders by predicting and denoising the entire latent sequence in parallel over a predicted duration; and 
(iii)~\textbf{Production-Hardening Post-Training}, utilizing a two-stage post-training recipe--a single-speaker voice lock that stabilizes speaker identity (collapsing cross-generation $F_0$ standard deviation from $74.9$\,Hz to $5.0$\,Hz) followed by short-utterance coverage to resolve isolated-token and short-phrase failures.
On our Freya-TR-Eval benchmark, \modelname{} achieves a band-matched WER of $8.0$\% and CER of $3.0$\%, lower error than both larger open systems in its field, XTTS-v2 and F5-TTS, at $40$--$55$\% of their parameter count, together with the highest naturalness (MOS) among the compact systems; two small phonemizer-driven VITS baselines retain lower WER at the cost of naturalness.
With a real-time factor of $0.11$ on a consumer GPU (RTX 4090; benchmark-set mean $\approx$$0.14$ on an H100) and real-time synthesis on a laptop CPU, the model is well suited to resource-constrained edge deployment.
We release the model weights, the training and inference code, and the evaluation benchmark under the Apache-2.0 license.
\end{abstract}

\clearpage
\tableofcontents
\clearpage

\section{Introduction}
\label{sec:intro}

Text-to-speech (TTS) has advanced from producing merely intelligible speech toward generating natural, expressive, and controllable audio~\citep{shen2018natural,renfastspeech}, driven by the large-language-model paradigm of framing synthesis as sequence modeling over discrete audio tokens~\citep{borsos2023audiolm,chen2025neural} and, more recently, by continuous-latent flow-matching generators~\citep{shen2023naturalspeech,le2023voicebox,chen2024f5}. This progress, however, is concentrated in high-resource English and Chinese. Mid-resource languages such as Turkish, for which public corpora such as FLEURS-tr~\citep{conneau2023fleurs} and Common-Voice-tr~\citep{ardila2020common} offer only tens of hours of validated read speech, remain underserved: large multilingual systems~\citep{casanova2024xtts,du2025cosyvoice3} treat Turkish as one language among dozens rather than as a first-class target. A second gap is computational. The strongest open systems reach their quality through multi-billion-parameter backbones trained on hundreds of thousands to millions of hours of speech (for example, CosyVoice\,3 on roughly one million hours and Qwen3-TTS on over five million)~\citep{du2025cosyvoice3,hu2026qwen3}, an operating point that excludes single-GPU serving and on-device deployment. Turkish additionally stresses the text frontend: agglutinative morphology, vowel harmony, and the spoken-form expansion of numbers, dates, and acronyms make hand-built grapheme-to-phoneme and normalization pipelines brittle and incomplete. These pressures argue for a compact, Turkish-first system that learns pronunciation end-to-end from audio rather than dictating it by rules.

Two modeling paradigms dominate contemporary TTS. Discrete-token systems represent speech as codec tokens and inherit LLM-style scaling and in-context learning~\citep{borsos2023audiolm,chen2025neural,du2025cosyvoice3}, but quantization discards fine acoustic detail and typically forces a multi-stage pipeline. Continuous-latent systems instead model speech representations directly with denoising or flow-matching objectives, spanning non-autoregressive diffusion models~\citep{shen2023naturalspeech,le2023voicebox,chen2024f5,eskimez2024e2} and diffusion-autoregressive hybrids~\citep{li2024autoregressive,jia2025ditar}. VoxCPM~\citep{zhou2025voxcpm} unified these lines through a differentiable finite-scalar-quantization bottleneck~\citep{mentzerfinite} inside a single continuous-latent backbone, and its successor VoxCPM2~\citep{zhou2026voxcpm2} scales the design to a 2B-parameter, 48\,kHz, 30-language foundation model whose asymmetric AudioVAE encodes at 16\,kHz and reconstructs at 48\,kHz. \modelname{} takes a different route: we treat its AudioVAE2 as a \emph{frozen}, Apache-2.0 continuous-latent codec, a fixed $25$\,Hz, $64$-dimensional latent space, and train a compact $183.2$M-parameter generator entirely inside it. Because high-fidelity $48$\,kHz reconstruction is already solved and held fixed, every parameter and training signal is spent on the text-conditional prior rather than on waveform modeling, and a from-scratch Turkish pretraining run over a large, high-quality corpus yields a strong Turkish-first model at a fraction of the size of the multilingual foundation models above.

The second design decision concerns \emph{how} the latent sequence is generated. VoxCPM and VoxCPM2 decode autoregressively, one acoustic patch at a time, atop a MiniCPM-4 backbone~\citep{team2025minicpm4}. Autoregression is expressive, but it accumulates errors over long horizons and is prone to number and word garbling, a benign glitch in a story reader but a critical failure wherever a misread digit changes meaning. \modelname{} therefore adopts a non-autoregressive formulation: a conditional flow-matching Diffusion Transformer predicts an entire utterance's latent sequence in parallel over a separately predicted duration, sidestepping left-to-right error accumulation by construction. Text conditions the latents through cross-attention over ConvNeXt-refined character features from a $92$-symbol Turkish character vocabulary, with no byte-pair encoding, phonemizer, or grapheme-to-phoneme step, so the model learns the pronunciation of in-context numbers, currencies, and acronyms directly from audio. What limits digit rendering in the NAR design is not decoding order but duration allocation, so digit strings are expanded to their spoken form at the text frontend, the standard input contract for character-level synthesis, while isolated bare tokens are handled by post-training and a thin inference layer (\cref{sec:method-posttrain,sec:exp}). A two-stage post-training recipe then specializes the from-scratch pretrained prior into a single-voice production model.

The main contributions of \modelname{} are as follows.

\begin{enumerate}
    \item \textbf{A tokenizer-free non-autoregressive Turkish TTS model.} \modelname{} is a 183.2M-parameter conditional flow-matching Diffusion Transformer that synthesizes 48\,kHz Turkish speech in the frozen continuous-latent space of AudioVAE2, from a 92-symbol character vocabulary with no phonemizer, grapheme-to-phoneme frontend, or discrete speech tokenizer,  to our knowledge the first openly released tokenizer-free NAR TTS model pretrained from scratch as a Turkish-first system, whereas prior open Turkish baselines relied on phonemizers or autoregressive architectures~\citep{pratap2024scaling,casanova2024xtts}. We release the single-voice production model's weights publicly.
    \item \textbf{A pretrain-to-SFT production-hardening recipe.} A full-parameter \emph{voice-lock} stage on a single consented speaker writes the voice into the weights, collapsing cross-generation $F_0$ standard deviation from $74.9$ to $5.0$\,Hz, and a \emph{short-utterance-coverage} stage over forced-aligned one-word and two-word segments repairs the isolated-token failures a sentence-only corpus structurally cannot.
    \item \textbf{A public benchmark and an efficiency-oriented evaluation.} We release \textbf{Freya-TR-Eval}, a \OneNitems{}-sentence, domain-neutral Turkish benchmark, together with its seeded build scripts and our full evaluation harness, so that every number in this report is reproducible without redistributing third-party audio. Under an identical band-matched Whisper WER/CER protocol, \modelname{} (WER \OneSftWer{}\,\%, CER \OneSftCer{}\,\%) attains lower error than both larger open systems in the comparison, XTTS-v2 and F5-TTS, and the highest MOS among the compact systems, while its batched non-autoregressive serving is markedly more throughput- and memory-efficient than the $2$B VoxCPM2 engine and runs faster than real time on one consumer GPU and in real time on a laptop CPU.
\end{enumerate}

The remainder of this report is organized as follows. \Cref{sec:related} situates \modelname{} within large-scale and low-resource TTS foundation models and continuous-latent flow-matching generation. \Cref{sec:method} presents the \modelname{} system: the tokenizer-free non-autoregressive architecture over the frozen VoxCPM2 latent space, the flow-matching and duration objectives, the from-scratch Turkish pretraining stage, and the two-stage voice-lock and short-utterance-coverage fine-tuning recipe. \Cref{sec:exp} reports the Turkish evaluation against openly available systems together with inference- and serving-efficiency measurements. \Cref{sec:conclusion} discusses limitations and future directions.

\section{Related Work}
\label{sec:related}
We situate \modelname{} along two axes: the generative paradigm it instantiates, non-autoregressive flow matching over a continuous latent space (\cref{sec:related-paradigm}), and the niche it fills, a compact Turkish-first model in a field of large multilingual generalists (\cref{sec:related-turkish}).

\subsection{Speech Synthesis Paradigms}
\label{sec:related-paradigm}
Contemporary TTS is organized by how speech is represented and generated. We review the three families that frame \modelname{}: discrete-token language models, non-autoregressive generation over continuous features, and diffusion-autoregressive hybrids.

\paragraph{Discrete-token codec language models.}
The dominant paradigm casts synthesis as autoregressive language modeling over discrete acoustic tokens from a neural audio codec~\citep{defossez2022high,kumar2023high}, inheriting the scaling and in-context learning of large language models. VALL-E~\citep{chen2025neural} cast zero-shot cloning as prompt continuation over residual-quantized tokens, and the CosyVoice family~\citep{du2024cosyvoice1,du2025cosyvoice3} pairs a language model over supervised semantic tokens with a separate flow-matching decoder that restores acoustic detail. The route scales well but pays two structural prices: quantization discards fine acoustic information, and the semantic-token and acoustic-decoder stages cannot be optimized jointly end to end.

\paragraph{Non-autoregressive generation over continuous features.}
A second family drops discrete tokens and models continuous acoustic features directly and in parallel, under a diffusion or conditional-flow-matching objective. NaturalSpeech\,2~\citep{shen2023naturalspeech} applies latent diffusion with explicit duration and pitch predictors, and Voicebox~\citep{le2023voicebox} frames TTS as text-conditioned flow matching over mel frames guided by an external alignment and a duration model. This alignment-and-duration template runs through diffusion TTS: Grad-TTS~\citep{popov2021grad} couples a score-based decoder with monotonic-alignment search and a duration predictor, and Matcha-TTS~\citep{mehta2024matcha} recasts it under optimal-transport conditional flow matching. Nearby, StyleTTS\,2~\citep{li2023styletts2} adds style diffusion and adversarial training, NaturalSpeech\,3~\citep{ju2024naturalspeech3} factorizes a codec latent for attribute-wise diffusion, and end-to-end VAE-GAN models of the VITS~\citep{kim2021conditional} family, on which the MMS Turkish baseline is built~\citep{pratap2024scaling}, learn alignment and waveform jointly. A distinct branch removes the alignment and duration modules altogether: E2-TTS~\citep{eskimez2024e2} pads the character sequence with filler tokens to the target length and lets a single Transformer learn the text-to-audio correspondence implicitly through self-attention, and F5-TTS~\citep{chen2024f5} keeps this filler-padded formulation while adding a ConvNeXt text refiner and sway-sampling inference; compact successors such as ZipVoice~\citep{zhu2025zipvoice} push the same objective toward small, fast models. \modelname{} belongs to this family but conditions through cross-attention over character features rather than filler-padded implicit alignment, and it generates a frozen VoxCPM2 latent rather than a mel-spectrogram.

\paragraph{Diffusion-autoregressive hybrids.}
A third family autoregresses over continuous latents while rendering each step with a local diffusion head, combining language-model-style planning with continuous acoustic fidelity~\citep{meng2024autoregressive}. The mechanism originates in image generation, where MAR~\citep{li2024autoregressive} replaced the categorical softmax of a token-based model with a small per-token diffusion loss, showing that vector quantization is not required for autoregressive generation. DiTAR~\citep{jia2025ditar} ported this to speech through a local diffusion transformer over patches of continuous latents, and VoxCPM~\citep{zhou2025voxcpm} realized a semantic-acoustic hierarchy inside a single continuous-latent backbone via a differentiable finite-scalar-quantization bottleneck~\citep{mentzerfinite} atop a MiniCPM-4 language model~\citep{team2025minicpm4}. These hybrids are expressive but retain an autoregressive decode loop, with the exposure bias and error accumulation it carries over long or number-dense utterances.

\paragraph{Positioning \modelname{}.}
\modelname{} keeps VoxCPM2's continuous-latent representation but discards its autoregressive stack. It reuses only the frozen AudioVAE, the $25$\,Hz, $64$-dimensional, $16$\,kHz-in / $48$\,kHz-out codec of VoxCPM2~\citep{zhou2026voxcpm2}, under its Apache-2.0 license and never trained, so representation learning is decoupled from generative modeling and $48$\,kHz super-resolution comes for free from the frozen decoder. In place of the hierarchical TSLM/FSQ/RALM stack and the autoregressive loop, a single $183.2$M-parameter flow-matching transformer predicts an utterance duration and denoises all latent frames in parallel (a $32$-step Euler ODE at inference). The system is tokenizer-free on both ends: no discrete audio codec, and a $92$-symbol character vocabulary with no phonemizer or grapheme-to-phoneme frontend, so that the pronunciation of numbers, acronyms, and code-switched terms is learned from audio rather than dictated by a rule-based frontend.

\subsection{Turkish, Multilingual, and Efficient TTS}
\label{sec:related-turkish}

\paragraph{The mid-resource regime: Turkish beyond nominal coverage.}
Progress in large-scale TTS is measured almost exclusively on English and Chinese. Training corpora are dominated by read and audiobook speech in those two languages, as in Emilia~\citep{he2024emilia}, and the zero-shot benchmarks that define the state of the art, notably Seed-TTS-Eval~\citep{anastassiou2024seed}, inherit the same distribution. Massively multilingual systems nominally widen this coverage to dozens or hundreds of languages~\citep{casanova2024xtts,du2025cosyvoice3,zhu2026omnivoice}, and Turkish typically appears among them. Nominal coverage, however, is not specialization: a single checkpoint must amortize a fixed capacity budget across every language it serves, and the supervision it receives for each is generic read speech. The consequence is sharpest for languages that are neither high- nor low-resource. Turkish occupies precisely this mid-resource band, where public read-speech corpora such as FLEURS-tr~\citep{conneau2023fleurs} and Common-Voice-tr~\citep{ardila2020common} together supply only tens of hours, an order of magnitude below their English or Chinese counterparts, and where no openly available corpus targets a conversational domain. A language in this regime is well served neither by the multilingual generalist, whose Turkish is a marginal slice of a shared budget, nor by the low-resource literature, whose methods assume a scarcity that does not hold. \modelname{} is designed for this regime: a Turkish-first model trained from scratch on a large in-domain Turkish corpus, rather than one language slot in a multilingual grid.

\paragraph{Compact and deployable synthesis.}
The strongest open systems obtain their quality from multi-billion-parameter backbones that preclude single-GPU serving and on-device use. A parallel line pursues efficiency instead: VITS-based on-device voices such as Piper and the multilingual MMS-TTS~\citep{pratap2024scaling}, and compact flow-matching models such as ZipVoice~\citep{zhu2025zipvoice}, trade breadth for a small, fast footprint. \modelname{} shares this goal but reaches it differently. A frozen high-fidelity codec absorbs waveform reconstruction, leaving a $183.2$M-parameter non-autoregressive generator that serves faster than real time on a consumer GPU and runs on a laptop CPU, while carrying a single production voice in its weights rather than cloning one from a reference prompt.

\section{Methodology}
\label{sec:method}

\subsection{Overview}
\label{sec:method-overview}

\begin{figure}[H]
\centering
\includegraphics[width=\textwidth]{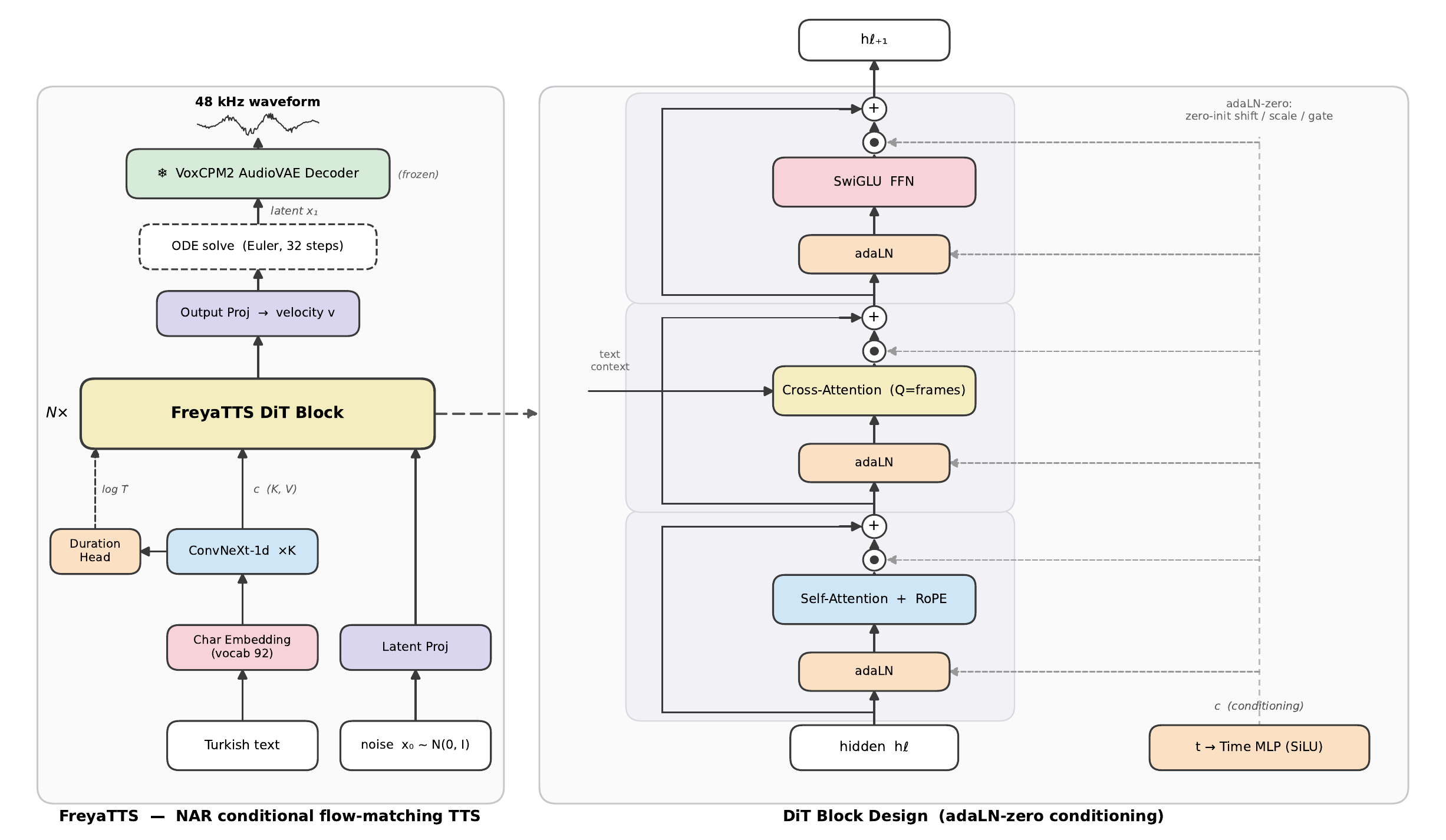}
\caption{Overall architecture of \modelname{}. Character-level Turkish text (a $92$-symbol vocabulary; no byte-pair encoding, phonemizer, or grapheme-to-phoneme frontend) is embedded and refined by a ConvNeXt-1d encoder into a character-feature sequence $c$, which (i)~drives a duration head that predicts the total latent length $\hat T$ and (ii)~serves as the key/value memory for the cross-attention layers of a non-autoregressive Diffusion Transformer (DiT). Conditioned on the flow-matching timestep $t$ through adaLN-zero modulation, the DiT denoises a length-$\hat T$ sequence of $64$-dimensional latent frames drawn from Gaussian noise; the resulting clean latents are rendered to $48$\,kHz audio by the \emph{frozen} AudioVAE2 decoder. The autoregressive TSLM/FSQ/RALM backbone is discarded; only AudioVAE2 is reused.}
\label{fig:arch}
\end{figure}

\modelname{} is a tokenizer-free, non-autoregressive (NAR) text-to-speech model realized as a conditional flow-matching Diffusion Transformer (DiT) over the continuous latent space of a \emph{frozen} AudioVAE2. It shares VoxCPM's core premise, modeling speech in a continuous latent space rather than over discrete codec tokens~\citep{zhou2025voxcpm}, but departs sharply from its generative backbone. VoxCPM and VoxCPM2 render audio autoregressively: a MiniCPM-4 language model~\citep{team2025minicpm4} predicts semi-discrete FSQ semantic states~\citep{mentzerfinite} that a local diffusion head decodes patch by patch. We instead discard the entire autoregressive stack (the text-semantic LM, the FSQ bottleneck, the residual acoustic LM, and the local diffusion transformer) and keep only AudioVAE2, replacing the backbone with a single NAR DiT that denoises all latent frames of an utterance in parallel (\cref{fig:arch}).

This choice is dictated by reliability. Conversational Turkish is dense with digit strings, dates, and amounts, and autoregressive latent prediction can accumulate error across such sequences, garbling numbers mid-utterance. Predicting the whole latent sequence in parallel removes the left-to-right dependency; it also decouples the number of \emph{sequential} generation steps from utterance length (a fixed ODE-step budget replaces a length-proportional decode loop, though each step still attends over all $T$ frames) and pairs naturally with flow matching~\citep{le2023voicebox,chen2024f5,anastassiou2024seed}, in contrast to the multi-stage LM-plus-vocoder pipelines of discrete-token systems~\citep{du2025cosyvoice3}.

\paragraph{Formulation.} Let $x_1\in\mathbb{R}^{T\times 64}$ be the clean latent sequence produced by the frozen AudioVAE2 encoder and $c=\mathrm{TextEnc}(y)$ the character-feature memory of the input text $y$. We adopt a conditional flow-matching objective with a linear probability path, which interpolates on a straight line between a Gaussian prior $x_0\sim\mathcal{N}(0,I)$ and the data $x_1$ and trains a velocity field $v_\theta$ to transport one to the other:
\begin{equation}
\label{eq:cfm}
\begin{aligned}
x_t &= (1-t)\,x_0 + t\,x_1, \qquad x_0\sim\mathcal{N}(0,I), \quad t\sim\mathcal{U}[0,1],\\[2pt]
\mathcal{L} &= \underbrace{\mathbb{E}_{t,\,x_0,\,x_1}\Big[\big\|\,m\odot\big(v_\theta(x_t,t,c)-\overbrace{(x_1-x_0)}^{v^\star}\big)\big\|_2^2\Big]}_{\text{masked flow-matching}}
\;+\;\lambda_{\mathrm{dur}}\,\big(\log\hat T-\log T\big)^2 .
\end{aligned}
\end{equation}
Here $v^\star=x_1-x_0$ is the target velocity, which is \emph{constant} along the straight path so that a single network evaluation supervises the entire trajectory; $m\in\{0,1\}^{T}$ masks padding frames, and the flow-matching term is averaged over the $\|m\|_1$ real frames; $\hat T$ is the length predicted by the duration head (\cref{sec:method-dit}); and $\lambda_{\mathrm{dur}}=0.1$. Because $x_0$ and $x_1$ are sampled independently rather than optimally coupled, this is the linear-path, \emph{rectified-flow} instance of conditional flow matching~\citep{lipman2023flow,tong2024improving}, not a minibatch optimal-transport coupling. There is no adversarial loss, no autoregression, and no discrete-token cross-entropy: only the masked flow-matching regression and the auxiliary duration term. Because the model consumes raw characters and no discrete speech tokenizer intervenes, the pronunciation of in-context numbers, currencies, and acronyms is learned end-to-end from audio rather than delegated to a text frontend; isolated tokens are harder and are addressed by post-training and inference-time guards (\cref{sec:method-posttrain}).

\subsection{Frozen AudioVAE2 Latent Space}
\label{sec:method-audiovae}

We build on AudioVAE2~\citep{zhou2026voxcpm2}, reused \emph{frozen} under its Apache-2.0 license and never trained. Its encoder maps a $16$\,kHz waveform to $64$-dimensional latent frames at $25$\,Hz (one frame per $40$\,ms) and its decoder reconstructs at a $48$\,kHz sample rate. We stress that this $16$\,kHz-in / $48$\,kHz-out asymmetry fixes the output \emph{sample rate}, not the acoustic bandwidth: the usable bandwidth of a synthesis is still bounded by the encoder's $16$\,kHz input and by the band of the training audio, so the narrowband single-speaker corpus of \cref{sec:method-posttrain} yields a $48$\,kHz-rate signal that retains a telephony-band fidelity ceiling (\cref{sec:conclusion}). Holding this codec fixed is what lets us train a compact generator from scratch: high-fidelity reconstruction is already solved, so all capacity goes to the text-conditional prior, and modeling the \emph{continuous} latent rather than discrete tokens keeps fine acoustic detail and leaves the pipeline tokenizer-free~\citep{shen2023naturalspeech,le2023voicebox}. We retain only this continuous codec; the semi-discrete FSQ bottleneck~\citep{mentzerfinite} that gives VoxCPM its semantic-acoustic factorization lives in the autoregressive backbone we discard, and our NAR DiT operates directly on the native $25$\,Hz frame grid with no patch grouping, so the sequence length $T$ is simply the number of $40$\,ms frames.

\subsection{The \modelname{} Diffusion Transformer}
\label{sec:method-dit}

The velocity field $v_\theta$ is a DiT that takes the noisy latent sequence $x_t\in\mathbb{R}^{T\times 64}$, the timestep $t$, and the character memory $c$, and predicts a velocity at every frame. \Cref{tab:config} lists the configuration.

\paragraph{Text encoder and duration head.} The text encoder embeds the character-level Turkish vocabulary ($92$ symbols in total: the $29$-letter alphabet including \c{c}, \u{g}, \i, \"o, \c{s}, \"u, together with digits, punctuation, and whitespace) and refines it with four ConvNeXt-1d blocks~\citep{chen2024f5,liu2023convnext} into the feature sequence $c$, shared by the cross-attention memory and the duration head. Operating on raw characters means that Turkish orthographic phenomena (vowel harmony, the dotted/dotless \i\ distinction, and the spoken form of acronyms) are absorbed into the weights rather than handed to a grapheme-to-phoneme frontend. Digit strings are the deliberate exception: their orthographic and phonetic lengths diverge so sharply that we expand them to spoken form at the text frontend, since the bottleneck lies in the duration predictor rather than in the acoustic model; isolated acronyms and bare tokens remain a documented failure mode addressed in \cref{sec:method-posttrain,sec:exp}. Since generation is non-autoregressive, the target length must be known before denoising: a small MLP over the mask-averaged mean of $c$ regresses $\log\hat T$, supervised by the auxiliary term of \cref{eq:cfm} and used to size the noise tensor at inference.

\paragraph{DiT block.} Each of the $16$ layers updates the frame hidden states $h\in\mathbb{R}^{T\times d}$ ($d{=}640$) through three residual sub-layers (RoPE self-attention, text cross-attention, and a SwiGLU feed-forward network), each modulated by the flow-matching timestep through adaLN-zero:
\begin{equation}
\label{eq:dit-block}
\begin{aligned}
(\beta_1,\gamma_1,g_1,\;\beta_2,\gamma_2,g_2,\;\beta_3,\gamma_3,g_3) &= \mathrm{MLP}_{\mathrm{zero}}\big(\mathrm{emb}(t)\big),\\
h &\leftarrow h + g_1\odot\mathrm{SelfAttn}_{\mathrm{RoPE}}\big(\gamma_1\odot\mathrm{LN}(h)+\beta_1\big),\\
h &\leftarrow h + g_2\odot\mathrm{CrossAttn}\big(\gamma_2\odot\mathrm{LN}(h)+\beta_2,\;c\big),\\
h &\leftarrow h + g_3\odot\mathrm{SwiGLU}\big(\gamma_3\odot\mathrm{LN}(h)+\beta_3\big),
\end{aligned}
\end{equation}
where $\mathrm{emb}(t)$ is a sinusoidal timestep embedding and $\odot$ applies per-channel shift/scale/gate broadcast over frames. The nine modulation vectors are produced by a per-layer MLP whose final projection is \emph{zero-initialized} (adaLN-zero): every block therefore begins as the identity and departs from it only as training warrants, a stabilizer standard in diffusion transformers~\citep{peebles2023scalable,li2024autoregressive,jia2025ditar}. Self-attention carries rotary position embeddings~\citep{su2024roformer} over the frame axis; in contrast to the NoPE~\citep{kazemnejad2023impact} residual LM of VoxCPM2~\citep{zhou2026voxcpm2}, we retain RoPE because our sequence is generated in one shot and benefits from explicit positional structure. In $\mathrm{CrossAttn}$ the queries are the modulated frame states while the keys and values are the character features $c$: this dedicated cross-attention is the sole pathway through which text conditions acoustics, following the alignment-based diffusion decoders of Grad-TTS, Matcha-TTS, and NaturalSpeech\,2~\citep{popov2021grad,mehta2024matcha,shen2023naturalspeech}.

\begin{table}[t]
\centering
\caption{\modelname{} configuration. AudioVAE2 is reused frozen; only the DiT and text encoder ($183.2$M parameters) are trained.}
\label{tab:config}
\small
\begin{tabular}{ll}
\toprule
\textbf{Component} & \textbf{Setting} \\
\midrule
Latent space (frozen AudioVAE2) & $64$-dim @ $25$\,Hz; $16$\,kHz encode / $48$\,kHz decode \\
Text vocabulary & character-level Turkish, $92$ symbols (no BPE, no G2P) \\
Text encoder & ConvNeXt-1d $\times\,4$ \\
DiT width / depth / heads & $d{=}640$ / $16$ / $10$ \\
DiT feed-forward & SwiGLU, hidden $2048$ \\
Self-attention position encoding & RoPE \\
Time conditioning & adaLN-zero ($9$-way modulation) \\
Text conditioning & cross-attention (frame queries $\to$ character keys/values) \\
Duration objective weight $\lambda_{\mathrm{dur}}$ & $0.1$ (pretraining) \\
Trainable parameters & $183.2$M \\
Inference solver & $32$-step Euler ODE \\
\bottomrule
\end{tabular}
\end{table}

\subsection{Pretraining from Scratch}
\label{sec:method-pretrain}

The DiT and text encoder are trained from random initialization (AudioVAE2 contributes no gradients) on a large-scale, high-quality internal corpus of multi-speaker Turkish speech, pre-encoded offline into AudioVAE2 latents so that the frozen encoder never runs during training, and spanning number-string reads, name-list reads, conversational sentences, and general speech. We optimize \cref{eq:cfm} in bf16 with AdamW (learning rate $5\!\times\!10^{-4}$, weight decay $0.01$, gradient clipping at $1.0$, $2{,}000$-step linear warmup then cosine decay) for $150$k steps at batch size $64$, in a single training run with a fixed seed. Training runs on H100 and H200 GPUs with a memory-safe loader over the pre-encoded fp16 latents.

\paragraph{Pretraining behavior.} The flow-matching loss descends smoothly to convergence over the $150$k steps (\cref{fig:loss}), and re-transcribing generations with Whisper-large-v3~\citep{radford2023robust} yields intelligible, well-aligned Turkish, with in-context numbers read in their full spoken form. The pretrained model has, however, \emph{no speaker identity}: with no speaker field in the corpus and no speaker conditioning in the network, the Gaussian prior $x_0$ seeds a fresh speaker on each generation. This is expected behavior for an unconditioned prior, and it directly motivates the voice-lock post-training below; \cref{sec:exp-voicelock} quantifies the effect.

\subsection{Post-training}
\label{sec:method-posttrain}

Two supervised fine-tuning stages turn the speaker-agnostic pretrained model into a deployable single-voice product.

\paragraph{SFT\,v1: single-speaker voice lock.} Initialized from the $150$k-step checkpoint, we full-parameter fine-tune (AudioVAE2 still frozen) on an internal single-speaker corpus of a single consented professional voice talent, recorded at $16$\,kHz over a narrowband channel matching the deployment domain. Training uses multi-GPU DDP, bf16, and learning rate $1\!\times\!10^{-4}$ with cosine decay; the flow-matching loss falls from $0.918$ to $\approx$$0.62$ within roughly one epoch. The speaker identity is written into the weights: cross-generation F0 standard deviation collapses from $74.9$ to $5.0$\,Hz and content similarity reaches $0.923$, while domain acronyms stabilize, pronounced as words rather than spelled letter by letter. We select \textbf{step\,1000}: continued training over-fits and destabilizes the voice, with the voice-lock standard deviation drifting from $5.0$ back to $11$\,Hz by $3$k steps.

\paragraph{SFT\,v2: short-utterance coverage.} The single-speaker corpus contains \emph{no} clips of two words or fewer, which we identified as the structural root cause of a collapse on isolated acknowledgments, bare numbers, and lone acronyms. We therefore continue fine-tuning from the v1 step-$1000$ checkpoint on the same corpus augmented with mined short single-speaker segments (one-word and two-word phrases, numbers, and acknowledgments) that we extracted from the same speaker by MMS forced alignment (torchaudio \texttt{MMS\_FA} with a Turkish-to-roman character map). This continuation stage uses learning rate $5\!\times\!10^{-5}$ and $\lambda_{\mathrm{dur}}{=}0.2$, and we again select step\,1000. Isolated acknowledgments are rendered once, cleanly, rather than entering a repetition loop; in-context numbers are correct; and full-sentence quality and the locked voice are preserved. \textbf{SFT\,v2 is the shipped production model.} A thin inference wrapper completes the system: clause-level chunking for long inputs, a duration floor for very short ones, and a voicing-based retry on the rare degenerate draw.

\subsection{Inference}
\label{sec:method-inference}

Given input text, the text encoder produces $c$ and the duration head predicts the latent length $\hat T$. We draw $x_0\sim\mathcal{N}(0,I)$ of length $\hat T$ and integrate the flow-matching ODE $\mathrm{d}x/\mathrm{d}t=v_\theta(x_t,t,c)$ from $t{=}0$ to $t{=}1$ with a fixed-step Euler solver ($32$ steps), evaluating the DiT once per step with $c$ supplied through cross-attention. The resulting clean latent sequence is decoded by the frozen AudioVAE2 into a $48$\,kHz waveform. Because generation is non-autoregressive, the whole sequence is denoised in parallel: the number of \emph{sequential} solver steps is fixed at $32$ regardless of utterance length (each step still attending over all $T$ frames), and there is no left-to-right error accumulation. Unlike the autoregressive sampler of VoxCPM2, which leans on classifier-free guidance, sway sampling~\citep{chen2024f5}, and CFG-Zero$^{*}$~\citep{fan2025cfgzerostar}, we find a plain $32$-step deterministic Euler integrator sufficient; the inference wrapper re-draws $x_0$ only on the rare voicing-check failure. Empirically the shipped \modelname{} model attains a mean per-utterance real-time factor of $\approx$$0.14$ on an H100, which we detail in \cref{sec:exp-efficiency}.

\begin{figure}[t]\centering
\begin{subfigure}{0.32\textwidth}\centering\includegraphics[width=\linewidth]{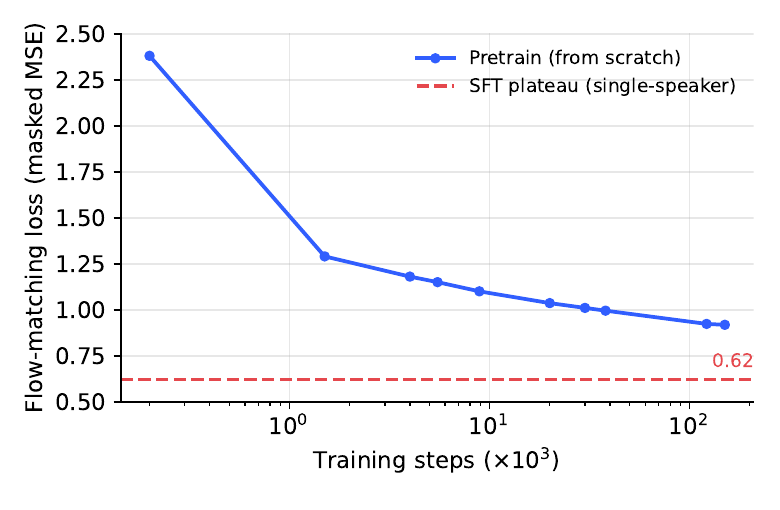}\caption{Training loss.}\label{fig:loss}\end{subfigure}\hfill
\begin{subfigure}{0.32\textwidth}\centering\includegraphics[width=\linewidth]{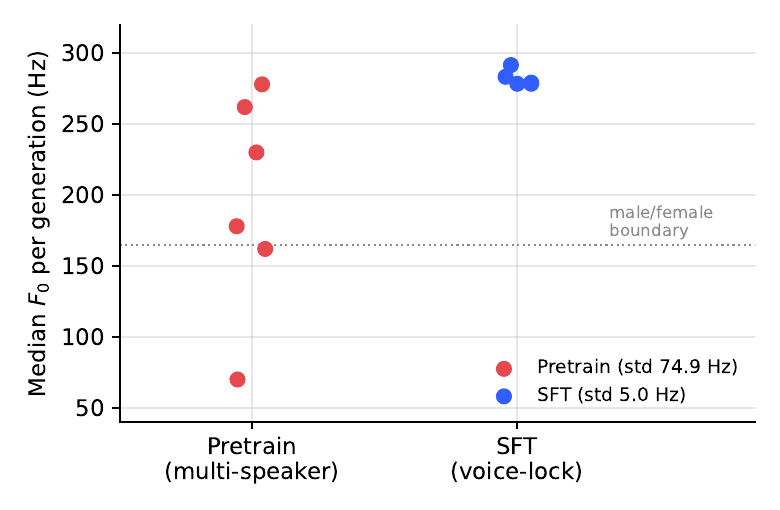}\caption{Voice-lock ($F_0$).}\label{fig:voicelock}\end{subfigure}\hfill
\begin{subfigure}{0.32\textwidth}\centering\includegraphics[width=\linewidth]{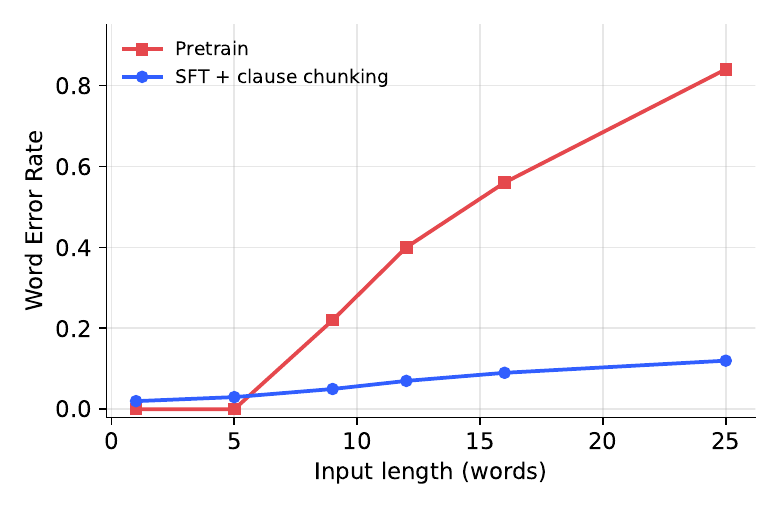}\caption{WER vs.\ length.}\label{fig:wer_length}\end{subfigure}
\caption{\textbf{Left:} flow-matching loss over pretraining (log-scale steps) with the single-speaker SFT plateau. \textbf{Center:} median $F_0$ across regenerations of a fixed prompt; the multi-speaker pretrained model scatters across an octave (std $74.9$\,Hz) whereas the voice-locked SFT model is stable (std $5.0$\,Hz). \textbf{Right:} word error rate as a function of input length on a length-swept probe set; long-horizon drift in the pretrained model is mitigated, though not removed, by inference-time clause chunking.}
\label{fig:results}
\end{figure}

\section{Experiments and Results}
\label{sec:exp}

We evaluate \modelname{} along three axes: main results against the field of openly-available, sub-billion-parameter Turkish TTS systems on a single held-out benchmark (\cref{sec:exp-setup,sec:exp-one}); speaker consistency and the voice-lock stage (\cref{sec:exp-voicelock}); and inference efficiency (\cref{sec:exp-efficiency}). 

\subsection{Experimental Setup}
\label{sec:exp-setup}

\paragraph{Benchmark.}
We evaluate on a single, purpose-built benchmark, \textbf{Freya-TR-Eval}: $495$ natural, domain-neutral
Turkish sentences of everyday conversational register (statements, \emph{mI}-questions, and exclamations),
length $3$--$13$ words. It merges native Turkish sentence texts sampled from Common Voice~17-tr
\citep{ardila2020common} and CoVoST2-tr \citep{wang2020covost} with everyday-conversational lines generated by
\texttt{gemini-3.1-pro-preview} under fixed topic and sentence-type quotas ($22$ everyday topics), filtered for
length, full Turkish grapheme coverage (ç\,ğ\,ı\,i\,ö\,ş\,ü), and de-duplication. \modelname{} is trained only
on internal corpora (\cref{sec:method-pretrain,sec:method-posttrain}) disjoint from the benchmark sources, so
every item probes generalization; the exact sentence list and the seeded build scripts are
released.\footnote{Released as \texttt{freyavoice/freya-tr-eval} on HuggingFace.} Because \modelname{} is a
single-voice synthesizer rather than a zero-shot cloner, we synthesize the reference text of each item and score
the output; the two reference-prompt baselines (XTTS-v2, F5-TTS) are given one fixed shared reference clip
(cloner scores can be sensitive to this choice).

\paragraph{Comparison systems.}
We position \modelname{}, the shipped $183.2$M single-voice model, against the field of
\emph{openly available, sub-billion-parameter} Turkish TTS systems, the models a practitioner can download
and self-host: \textbf{MMS-TTS-tr}~\citep{pratap2024scaling} ($36$M, Meta's VITS-based multilingual TTS),
\textbf{Coqui GlowTTS-tr} ($\sim$$28$M, flow-based), \textbf{Piper-tr} ($\sim$$16$M, VITS, on-device),
\textbf{SpeechT5-tr}~\citep{ao2022speecht5} ($\sim$$144$M, fine-tuned on Turkish Common Voice),
\textbf{XTTS-v2}~\citep{casanova2024xtts} ($\sim$$470$M, multilingual zero-shot cloner), and
\textbf{F5-TTS-Turkish}~\citep{chen2024f5} ($\sim$$336$M, a flow-matching zero-shot cloner fine-tuned for Turkish).

\paragraph{Metrics.}
The primary metrics are \textbf{band-matched WER and CER}: every system's output is downsampled to $8$\,kHz
before transcription with Whisper-large-v3~\citep{radford2023robust}, so that \modelname{}'s telephony-band
$48$\,kHz output (\cref{sec:method-audiovae}) is not penalized relative to natively wideband baselines, and
reference and transcript pass the same Turkish text normalization before scoring; the protocol is applied
identically to every system. We also report
\textbf{real-time factor} (RTF): per-utterance synthesis time over audio duration at batch size $1$, including
AudioVAE2 decoding, averaged over the set, on an H100. Naturalness is measured with a \textbf{mean opinion score} (MOS) listening study: $24$ native Turkish raters scored randomized, system-blind samples of every system on a $5$-point scale ($2566$ ratings in total, $360$--$375$ per system); we report per-system means with $95\%$ confidence intervals. \textbf{Voice consistency}
(\cref{sec:exp-voicelock}) is the $F_0$ standard deviation across regenerations of the same text.
\textbf{Content similarity} is $1$ minus the normalized character-level edit distance between the input text
and the Whisper-large-v3 transcript of the synthesis.

\subsection{Main Results}
\label{sec:exp-one}

\begin{table}[H]\centering
\caption{\textbf{Conversational Turkish TTS, sub-1B open models.} All systems evaluated on the full Freya-TR-Eval set ($495$ everyday, general-purpose Turkish sentences) under an identical protocol: \textbf{WER}/\textbf{CER} from Whisper-large-v3 with $8$\,kHz band-matching (in \%), and naturalness \textbf{MOS} from a listening study with $24$ native Turkish raters ($2566$ ratings, $360$--$375$ per system, $5$-point scale, $95\%$ confidence intervals). Best per column in \textbf{bold}; systems ordered by size.}
\label{tab:one}
\setlength{\tabcolsep}{8pt}\small
\begin{tabular}{lcccc}
\toprule
\textbf{System} & \textbf{Params} & \textbf{WER $\downarrow$} & \textbf{CER $\downarrow$} & \textbf{MOS $\uparrow$} \\
\midrule
Piper (tr, dfki) & ~16M & \textbf{4.4} & \textbf{1.1} & $3.47 \pm 0.22$ \\
Coqui GlowTTS (tr) & ~28M & 12.1 & 3.3 & $2.53 \pm 0.19$ \\
MMS-TTS (tr) & 36M & 6.8 & 1.7 & $3.58 \pm 0.20$ \\
SpeechT5 (tr) & ~144M & 83.4 & 45.5 & $1.59 \pm 0.14$ \\
\textbf{FreyaTTS (ours)} & 183.2M & 8.0 & 3.0 & $3.68 \pm 0.22$ \\
\midrule
F5-TTS (tr) & ~336M & 24.3 & 10.9 & $3.63 \pm 0.23$ \\
XTTS-v2 (multi) & ~470M & 11.1 & 3.9 & $\textbf{3.82} \pm 0.19$ \\
\bottomrule
\end{tabular}
\end{table}

\paragraph{\modelname{} attains lower error than the larger open systems in its field.}
\Cref{tab:one} reports band-matched Whisper WER/CER for every system on the same $\OneNitems{}$
conversational sentences. \modelname{} reaches WER \OneSftWer{}\,\% and CER \OneSftCer{}\,\%,
\textbf{lower error than both larger open systems in the field},
the zero-shot cloners XTTS-v2 ($\sim$$470$M, WER \OneXttsWer{}\,\%) and F5-TTS ($\sim$$336$M, WER
\OneFfiveWer{}\,\%), each $1.8$--$2.6\times$ its parameter count, despite their larger backbones and far greater
training data: cloning a target voice from a short reference generalizes poorly to arbitrary sentences, whereas
\modelname{} carries a single voice natively in its weights. It also comfortably beats the weaker same-class
models (GlowTTS, SpeechT5). Piper ($\sim$$16$M, WER \OnePiperWer{}\,\%) and MMS-TTS ($\sim$$36$M, WER
\OneMmsWer{}\,\%) are compact phonemizer-driven VITS baselines whose highly regular pronunciation maximizes
recognizer agreement; \modelname{} supplies the single, consistent \emph{learned} voice they lack
(\cref{sec:exp-voicelock}). The listening study bears this out on naturalness: \modelname{} reaches the second-highest MOS point estimate ($3.68$), above the two low-WER VITS baselines (Piper $3.47$, MMS-TTS $3.58$) and F5-TTS ($3.63$) and behind only the $2.6\times$-larger XTTS-v2 ($3.82$); the $95\%$ intervals of the top four systems overlap, so we read the MOS column as a ranking of point estimates rather than pairwise significance. That a $183$M model trained from scratch attains lower WER/CER than both larger open systems in this field, while ranking second on naturalness, is the
paper's central empirical claim.

\paragraph{Statistical robustness, calibration, and the inference layer.}
Three checks contextualize the headline numbers. First, a sentence-level bootstrap ($10{,}000$ resamples over the $495$ items, computed on an independent re-synthesis of the full set) gives WER $7.9\%$ with a $95\%$ confidence interval of $[6.8, 9.1]$ and CER $2.9\%$ $[2.4, 3.4]$, reproducing \cref{tab:one} within run-to-run noise. Second, two anchors calibrate the protocol's absolute scale: real human recordings (FLEURS-tr test~\citep{conneau2023fleurs}) transcribe at $9.7\%$ WER under the identical band-matched pipeline, and re-encoding held-out \emph{real} recordings of the target speaker through the frozen AudioVAE alone costs $2.6\%$ WER -- the transcription floor of the latent space itself. \modelname{}'s $8.0\%$ is therefore comparable in magnitude to the recognizer's error on real human speech, though the two are measured on different corpora and Whisper WER is sensitive to text distribution, and sits within $5.4$ points of the latent space's floor. Third, ablating the entire inference layer of \cref{sec:method-inference} (normalization, clause chunking, duration floor, voicing retry) moves WER by at most $0.3$ points (paired bootstrap $95\%$ CI of the difference $[-0.3, 0.0]$); the voicing retry fires on $1$ of $495$ sentences and clause chunking on $7$ of $495$. The guards are tail insurance against rare failure modes, not a crutch the headline numbers depend on. Error does climb with input length on a length-swept probe (\cref{fig:wer_length}): long-horizon drift on long utterances is mitigated, though not eliminated, by this clause chunking.

\subsection{Speaker Consistency and the Voice-Lock}
\label{sec:exp-voicelock}

Because \modelname{} carries no speaker field and no reference-prompt pathway, the pretrained prior has no notion of speaker identity: the Gaussian prior $x_0$ seeds a fresh, uncontrolled speaker on every generation. \Cref{fig:voicelock} makes this concrete. Regenerating a fixed text many times (a $114$-generation voice-consistency probe), the pretrained model's per-generation median $F_0$ spans roughly $70$--$347$\,Hz, with a standard deviation of $74.9$\,Hz and a $60\%$ rate of crossing the $165$\,Hz male/female pitch boundary between successive draws. Intelligibility metrics are blind to this (the content is correct on each draw), yet for a deployed single-voice product it is a categorical failure: the listener hears a different person each turn.

The voice-lock stage writes a single identity into the weights. Full-parameter fine-tuning on the consented target speaker (SFT-v1; \cref{sec:method-posttrain}) collapses the cross-generation $F_0$ standard deviation from $74.9$\,Hz to $5.0$\,Hz and effectively eliminates the gender flips, while the content-similarity metric of \cref{sec:exp-setup} reaches $0.923$: the model now emits a deterministic, consistent target voice regardless of the noise draw (\cref{fig:voicelock}), measured on sentence-length in-distribution text. A speaker-verification embedding confirms the lock at the identity level: the ECAPA-TDNN~\citep{desplanques2020ecapa} cosine similarity between \modelname{} syntheses and the target-speaker centroid is $0.873 \pm 0.029$, within one standard deviation of the $0.883 \pm 0.026$ same-speaker ceiling measured on disjoint \emph{real} recordings of the speaker, and far above the $0.140$ different-speaker floor. The identity is a property of the parameters, not of a prompt: no reference audio is supplied at inference. We select the step-$1000$ checkpoint because prolonged fine-tuning over-specializes and destabilizes the voice, with the consistency standard deviation drifting back from $5.0$\,Hz toward $11$\,Hz by step $3$k. SFT also stabilizes pronunciation: all-caps acronyms are spoken natively as words rather than spelled out, and isolated short utterances, covered by SFT-v2's mined one-word and two-word segments (\cref{sec:method-posttrain}), are rendered cleanly. This pretrain-to-SFT contrast, an unconditioned multi-speaker prior turned into a locked single voice, is the mechanism by which the foundation-style generative prior becomes a deployable single-voice product.

\subsection{Inference Efficiency}
\label{sec:exp-efficiency}

\modelname{} generates $48$\,kHz audio non-autoregressively: given the duration head's predicted length $\hat T$, a single fixed-step Euler integrator runs $32$ ODE steps from noise to a clean latent sequence, which the frozen AudioVAE decodes in one pass (\cref{sec:method-inference}). The number of \emph{sequential} solver steps is fixed at $32$ regardless of $\hat T$, there is no left-to-right error accumulation, and a plain deterministic integrator suffices, without classifier-free guidance or sway sampling. At $183.2$M parameters the model fits in $1.5$\,GB of VRAM.

\begin{table}[!t]\centering
\caption{\textbf{Inference efficiency on a single RTX 4090} (fp32/bf16 as released, batch size $1$, end-to-end including vocoder/VAE decoding; median of $10$ runs after $3$ warmups on a fixed Turkish prompt set of three length buckets (five sentences each; released with our code)). \textbf{TTFT} = time to first audio on the long bucket ($22$--$28$ words): native streaming first-chunk for XTTS-v2, first synthesized clause for \modelname{}, full-utterance latency for the offline systems. \textbf{RTF} = synthesis time over audio duration. \textbf{Throughput} = aggregate audio-seconds per wall-second over a $45$-utterance queue at the best concurrency level ($C$ engines in separate processes). Best per column in \textbf{bold}. $^{\dagger}$Serving-style rows: VoxCPM2 through the nanovllm continuous-batching engine (chunk streaming, $16$ concurrent sequences, GPU-memory utilization $0.5$), and \modelname{} through a static-batching ODE server (padded batch, one masked $32$-step solve; TTFT column reports its per-batch latency at $B{=}8$). All other rows are the systems' released library stacks.}
\label{tab:speed}
\setlength{\tabcolsep}{5pt}\small
\begin{tabular}{lcccccc}
\toprule
\textbf{System} & \textbf{Params} & \textbf{VRAM (GB)} & \textbf{TTFT$_{\text{long}}$ (s)} & \textbf{RTF$_{\text{med}}$} & \textbf{RTF$_{\text{long}}$} & \textbf{Throughput} \\
\midrule
\textbf{\modelname{} (ours)} & 183M & 1.5 & 0.52 & \textbf{0.11} & 0.10 & 9.4 ($C{=}4$) \\
F5-TTS (tr) & $\sim$336M & \textbf{0.8} & 0.74 & 0.13 & \textbf{0.05} & 8.4 ($C{=}2$) \\
XTTS-v2 (multi) & $\sim$470M & 2.2 & 0.30 & 0.33 & 0.33 & 3.4 ($C{=}2$) \\
Spark-TTS & $\sim$0.5B & 5.4 & 14.0 & 1.08 & 1.04 & 1.4 ($C{=}2$) \\
VoxCPM2 & $\sim$2B & 5.5 & 4.2 & 0.37 & 0.37 & 1.7 ($C{=}1$) \\
\midrule
VoxCPM2 (serving engine)$^{\dagger}$ & $\sim$2B & 11.4 & \textbf{0.14} & 0.15 & 0.13 & 24.8 ($C{=}16$) \\
\textbf{\modelname{} (batched ODE)}$^{\dagger}$ & 183M & 3.8 & 0.54 & -- & -- & \textbf{65.2} ($B{=}8$) \\
\bottomrule
\end{tabular}
\end{table}

\Cref{tab:speed} measures \modelname{} against four larger open systems, including the $\sim$$2$B VoxCPM2~\citep{zhou2026voxcpm2} whose AudioVAE2 we reuse, the autoregressive Spark-TTS~\citep{wang2025spark}, XTTS-v2~\citep{casanova2024xtts}, and F5-TTS~\citep{chen2024f5}, each run end-to-end through its released inference stack with default sampling on an RTX 4090. Among the released library stacks, \modelname{} attains the best medium-sentence RTF ($0.11$) and the highest throughput ($9.4$ audio-seconds per wall-second with four concurrent engines) at the second-smallest memory footprint. Against the production-class VoxCPM2 it is $3.4\times$ faster in medium-bucket RTF, $8\times$ faster to first audio on long inputs, $3.7\times$ smaller in VRAM, and $5.5\times$ higher in throughput; against the autoregressive Spark-TTS the gaps widen to roughly $10\times$ in RTF and $27\times$ in TTFT. The optimized F5-TTS runtime is the one system in the same speed class: its fused bf16 kernel path reaches a lower long-form RTF, while \modelname{}, running unoptimized fp32 eager PyTorch, retains lower TTFT, higher throughput, and, per \cref{tab:one}, a three-times-lower Turkish WER with no reference clip at inference. XTTS-v2's native chunk streaming yields the best library-stack TTFT ($0.30$\,s) at roughly $3\times$ the RTF. We additionally measure VoxCPM2 through a dedicated continuous-batching serving engine (nanovllm) rather than its released library stack: chunk streaming brings its TTFT to $0.14$\,s and batching lifts throughput to $24.8$ audio-seconds per wall-second at $16$ concurrent sequences, at $11.4$\,GB of reserved VRAM. A serving engine is thus worth roughly an order of magnitude on both axes for the $2$B model. The same treatment favors the non-autoregressive design even more: a static-batching ODE server for \modelname{} (pad the batch, run one masked $32$-step solve, decode, trim; released with our inference tooling) reaches $65.2$ audio-seconds per wall-second at batch size $8$ with a $0.54$\,s per-batch latency and $3.8$\,GB of VRAM, $2.6\times$ the $2$B engine's continuous-batching throughput at one-third of its memory, before any length-bucketing or chunk-streaming refinements. On an H100, \modelname{}'s benchmark-set mean RTF is $\approx$$0.14$.

The compact size also admits deployment classes the systems of \cref{tab:speed} do not target. On a laptop CPU (Apple M3, four threads, fp32, no quantization) \modelname{} synthesizes in real time: RTF $0.70$ on medium-length sentences and $0.94$--$1.00$ on the short and long buckets, with a $5.7$\,s first-clause TTFT on long inputs. Compiled to Core ML at fp16, the DiT's full $32$-step loop renders $4.4$\,s of audio in $0.205$\,s on the laptop's neural engine (RTF $0.047$) and $0.72$\,s on CPU only (RTF $0.165$), with AudioVAE decoding adding $0.33$\,s on CPU, for an end-to-end real-time factor of $\approx$$0.12$ on consumer Apple silicon. A model competitive with the fastest GPU systems of its field while synthesizing in real time on a laptop CPU is an operating point that pairs the foundation-style generative prior with genuinely edge-deployable inference.

\section{Conclusion and Future Work}
\label{sec:conclusion}

In this work, we presented \modelname{}, a tokenizer-free, non-autoregressive flow-matching model for Turkish conversational speech synthesis. Built as a compact $183.2$M-parameter conditional flow-matching Diffusion Transformer inside the \emph{frozen} $25$\,Hz continuous-latent space of the AudioVAE2~\citep{zhou2026voxcpm2}, and driven by a $92$-symbol character vocabulary with no phonemizer, grapheme-to-phoneme frontend, or discrete speech tokenizer, \modelname{} is trained from scratch on a large-scale Turkish speech corpus and denoises an entire utterance in parallel in a fixed $32$-step ODE. A two-stage pretrain-to-SFT recipe, a single-speaker voice lock ($F_0$ standard deviation $74.9\!\rightarrow\!5.0$\,Hz) followed by short-utterance coverage, converts the speaker-agnostic prior into a reliability-hardened single-voice production model. Evaluated under a band-matched Whisper WER/CER protocol on the released \textbf{Freya-TR-Eval} benchmark, \modelname{} (WER \OneSftWer{}\,\%, CER \OneSftCer{}\,\%) attains lower error than both larger open systems, XTTS-v2 and F5-TTS, at a benchmark-set mean real-time factor of $\approx$$0.14$ on an H100 and roughly $40$--$55\%$ of their parameter count.

While \modelname{} delivers strong results for its size, several challenges remain. Two compact phonemizer-driven VITS baselines still attain lower error on clean conversational text; digit-dense input requires spoken-form expansion at the text frontend, because the character-level duration predictor under-allocates frames for compact digit strings; and the shipped voice inherits a narrowband fidelity ceiling from its telephony-band source. Future work will focus on a jointly learned CTC-monotonic aligner, already prototyped, that targets the residual word-skip and long-horizon-drift failures a single global length prediction cannot prevent; preference-based post-training~\citep{rafailov2023direct} from intelligibility and voicing rewards; extending the voice-lock recipe from one speaker to multiple voices and natural-language voice design; and explicit bandwidth extension so that a narrowband-trained speaker can be rendered in true wideband. We hope the open release of \modelname{}, the model weights, the Freya-TR-Eval benchmark, and the inference tooling provides a solid foundation for Turkish and low-resource speech research.

\section*{Contributors}
Ahmet Erdem Pamuk, Ömer Yentür, Ahmet Tunga Bayrak, Yavuz Alp Sencer Öztürk, Mustafa Yavuz.

\bibliographystyle{citation}
\bibliography{citation}

\end{document}